 \documentclass[letterpaper, 10 pt, conference]{ieeeconf}
\usepackage{todonotes}
\usepackage{amsmath}
\usepackage{graphics}
\usepackage{microtype}  
\usepackage{algorithm}
\usepackage{algpseudocode}

\usepackage{subcaption}
\usepackage{graphicx}

\usepackage{array}  
\usepackage{graphicx}  
\usepackage{amsmath, bm}  

\usepackage{amsthm}   
\usepackage{amsfonts} 
\usepackage{amssymb} 
\usepackage{mathtools}
\usepackage{glossaries}
\let\labelindent\relax

\usepackage{enumitem}
\usepackage{xcolor}
\usepackage{import}

\usepackage{cite}

\IEEEoverridecommandlockouts  

\glsdisablehyper
\newacronym{ml}{ML}{Machine Learning}
\newacronym{rl}{RL}{Reinforcement Learning}
\newacronym{mdp}{MDP}{Markov Decision Process}

\newacronym{reps}{REPS}{Relative Entropy Policy Search}

\newacronym{mp}{MP}{Movement Primitive}
\newacronym{dmp}{DMP}{Dynamic Movement Primitive}
\newacronym{codmp}{CoDMP}{Cooperative Dynamic Movement Primitive}
\newacronym{cmp}{CMP}{Compliant Movement Primitive}
\newacronym{promp}{ProMP}{Probabilistic Movement Primitive}
\newacronym{kmp}{KMP}{Kernelized Movement Primitive}

\newacronym{ilc}{ILC}{Iterative Learning Control}

\newacronym{gmmr}{GMM/R}{Gaussian Mixture Model/Regression}

\newacronym{rbf}{RBF}{Radial Basis Function}
\newacronym{gmm}{GMM}{Gaussian Mixture Model}
\newacronym{nf}{NF}{Normalizing Flow}

\newacronym{qcqp}{QCQP}{Quadratically Constrained Quadratic Program}

\newacronym{dof}{DOF}{Degree of Freedom}

\global\long\def\mymatrix#1{\boldsymbol{#1}}%

\global\long\def\myvec#1{\boldsymbol{#1}}%

\newcommand{\realset}{\mathbb{R}}
\newcommand{\symgroup}[1]{{\boldsymbol{\mathcal{S}}}^{#1}} 
\newcommand{\sogroup}{\mathrm{SO(3)}} 
\newcommand{\spdmanifold}[1]{{\boldsymbol{\mathcal{S}}}_{++}^{#1}} 
\newcommand{\spdgroup}[1]{{\boldsymbol{\mathcal{S}}}_{++}^{#1}}

\newcommand{\tangentspace}{\mathcal{T}}

\newcommand{\sourcepoint}{{\mymatrix{S}_i}}
\newcommand{\sourcemean}{{\overline{\mymatrix{S}}}}
\newcommand{\sourcepointtilde}{{\mymatrix{S}_i^{rct}}}

\newcommand{\targetmean}{{\overline{\mymatrix{T}}}}
\newcommand{\targetpoint}{{\mymatrix{T}_i}}

\newcommand{\targetpointtilde}{{\mymatrix{T}_i^{rct}}}

\renewcommand{\baselinestretch}{0.97}

\begin{document}
%

\title{Human-to-Robot Manipulability Domain Adaptation with Parallel Transport and Manifold-Aware ICP}

\author{Anna Reithmeir, Luis Figueredo, and Sami Haddadin
\thanks{
This work was partially funded by the Lighthouse Initiative Geriatronics by StMWi Bayern (grant 5140951), LongLeif GaPa gGmbH (grant 5140953), by the BMBF Projects 16SV7985 (KoBo34) AND the Deutsche Forschungsgemeinschaft (DFG) as part of EXC 2050/1 – Project 390696704 – Cluster of Excellence “Centre for Tactile Internet with Human-in-the-Loop” (CeTI).
Please note that S. Haddadin has a potential conflict of interest as shareholder of Franka Emika GmbH.}
\thanks{Authors are with the Munich Institute of Robotics and Machine Intelligence (MIRMI), Technische Universitat Munchen (TUM), 80637, Munich, Germany.  Email:\texttt{\{anna.reithmeir, luis.figueredo,  haddadin\}@tum.de}}}


%


\maketitle

\begin{abstract}

Manipulability ellipsoids efficiently capture the human pose and reveal information about the task at hand. 
Their use in task-dependent robot teaching -- particularly their transfer from a teacher to a learner -- can advance emulation of human-like motion.  
Although in recent literature focus is shifted towards manipulability transfer between two robots, the adaptation to the capabilities of the other kinematic system is to date not addressed and research in transfer from human to robot is still in its infancy.
This work presents a novel manipulability domain adaptation method for the transfer of manipulability information to the domain of another kinematic system. As manipulability matrices/ellipsoids are symmetric positive-definite (SPD) they can be viewed as points on the Riemannian manifold of SPD matrices. We are the first to address the problem of manipulability transfer from the perspective of point cloud registration. 
We propose a manifold-aware Iterative Closest Point algorithm (ICP) with parallel transport initialization.
Furthermore, we introduce a correspondence matching heuristic for manipulability ellipsoids based on inherent geometric features.
We confirm our method in simulation experiments with 2-DoF manipulators as well as 7-DoF models representing the human-arm kinematics. 









%
%

\end{abstract}


%
\IEEEpeerreviewmaketitle

\section{Introduction}

In manufacturing activities and daily living, humans are capable of performing complex tasks in a rather intuitive manner. In recent years, robotics research has given an increasing attention to the learning and
emulation of such human capabilities when performing elaborate manipulation tasks.

Learning by demonstration (LbD), i.e. learning modular representations as primitives from human demonstrations, offers a convenient way to embed task constraints and allow end-users to teach complex manipulation tasks to the robot. 
%
Notwithstanding, existing solutions mainly rely on kinesthetic teaching whereas -- while physically demonstrating a task to a robot -- humans often do not fully understand the robot's posture and physical constraints. Indeed, they are primarily concerned with moving the robot's end-effector in task space without reaching joint limits. 
Similar behavior can be observed with the integration of motion-capture systems. As a consequence, the human's preferences and cognition towards the task at hand, as well as their  
embodiment-based constraints and posture adaptation, are hindered. 
Resting on this observation, this work addresses the challenges of 
transferring human task specific posture -- together with its implicit manipulation capabilities -- 
to a robotic system whilst explicitly taking into account the differences in human-to-robot embodied kinematic structures.

\begin{figure}[t]
    \centering
    \includegraphics[width=\columnwidth]{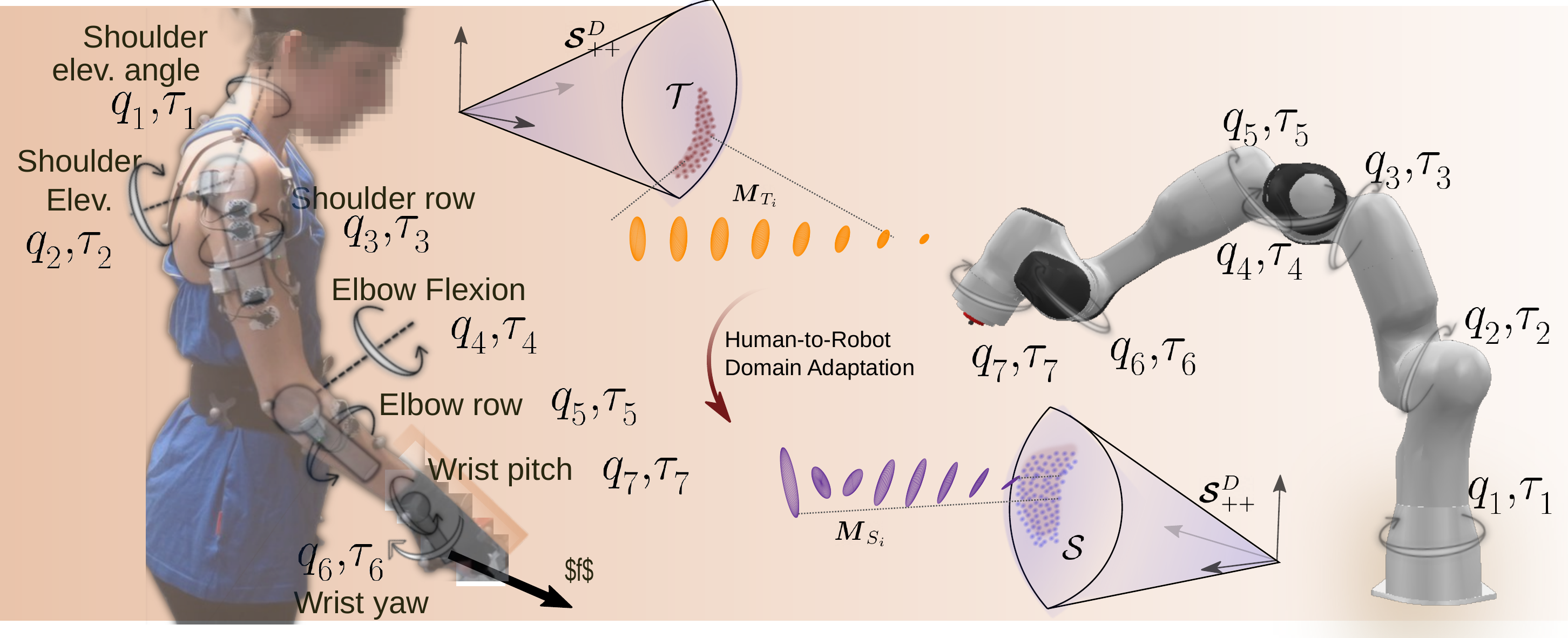}
    \caption{\small 
    The two 7-DoF models of a human and a robot arm give rise
    to distinct manipulability domains $\mathcal{T}$ and $\mathcal{S}$ on the SPD manifold, denoted as $\spdmanifold{D}$. The human arm is represented by the musculoskeletal model of \cite{Saul2015}. Human manipulabilities are transferred to
    the Franka Emika Panda robot by means of domain adaptation on the
    manifold.
    }
    \label{fig:big_picture}
\end{figure}

When performing manipulation tasks, humans exploit their cognitive understanding about the task to improve their balance, obstacle and joint-limit avoidance, energy-minimization, and posture  \cite{2009_Khatib_Demircan_De_Sapio_Sentis_Besier_Delp, figueredo2021planning_THRI}. Particularly with respect to the arm posture, it is well known that the human central nervous system adapts the arm configuration in accordance with the directional sensitivity of the task \cite{sabes1997obstacle}.  
In other words, human manipulation is characterized not only by the end-effector trajectory but 
also the arm posture, which is intrinsically captured in terms of the kinematic manipulability. More specifically, 
manipulability reveals the 
capability of the robot to apply force and motion in the different directions of task space 
given the current joint configuration by means of the task Jacobian. This relation is more explicitly depicted by the $L_2$-norm projection which gives rise to the manipulability ellipsoids  \cite{1985_Yoshikawa_ICRA,1986_Nakamura_Hanafusa_JDSMC,1988_Khatib_ISRR,2015_Patel_Tarek_JIRS,jaquier}. 
%

%

In terms of human manipulation,
manipulability analysis has successfully revealed the human preferences and cognition during walking  \cite{walking}, industrial-like scenarios \cite{jaquieranalysis} and reaching tasks \cite{biomechanicdecisions}.  
This has been extended to consider biomechanic factors and their influence on task dependent motion planning \cite{biomechanicdecisions,figueredo2020human}, safety-perception and comfortability features for physical human-robot interaction (pHRI) \cite{2018_Lipeng_Humanoids}, as well as to provide a major axis correlation with the direction of task dependent accelerations to minimize musculoskeletal effort \cite{2009_Khatib_Demircan_De_Sapio_Sentis_Besier_Delp}. 
Indeed, 
Jaquier et al. \cite{jaquieranalysis} have shown that the human posture and its manipulability adapt to the currently executed task to optimally fulfill the task constraints. For example, when carrying a heavy object the major axis of the manipulability ellipsoid aligns with the vertical task space axis to generate larger forces with fewer joint torques. Furthermore, in \cite{jaquieranalysis}, it is shown that the manipulabilities also reveal information about the intended task during the task preparation stage -- highlighting the importance of the manipulability information throughout a trajectory.  







When working with manipulability ellipsoids it is crucial to keep their 
underlying geometry in mind. Their topology results from the projection of the unit sphere within the joint-space (i.e., the $L_2$-norm of the torque or velocities) to task space variables through the local geometric Jacobian mapping -- yielding ellipsoid elements which are inherent symmetric positive-definite  (SPD) matrices. The set of SPD elements is defined within a cone manifold and is endowed with a Riemannian metric. This feature has long been overlooked by roboticists, which hindered several applications in the context of LbD until very recently \cite{geometryawaredmps,jaquier}.   
%
%
%
%
Notwithstanding, applications for manipulability learning in the context of LbD, as in the excellent work of Jaquier et al. \cite{jaquier}, are limited to a single SPD domain that reflects training and execution within the same kinematic structure, i.e., within the same robot.  

For the reproduction of human-like motion and cognitive understanding of a given task, instead, 
a robot would need to learn and emulate the human 
manipulability carried out throughout the task. 
This gives rise to a new challenge: The 
kinematic structure of the robot model can differ significantly from the one 
of the human body, and so do its capabilities.  
A similar problem concerning manipulability adaptation has, in fact, already been proposed as an open problem in \cite{jaquier}, but to this date not been addressed.  

In this work, we propose for the first time a transfer strategy which adapts the manipulability ellipsoid to exploit the capabilities of the learner.
Particularly focused on human-to-robot demonstrations, 
our method explores the differences between the manipulability domains --
i.e. the set of achievable manipulability ellipsoids -- of the human and the robot in order to find a locally optimal transformation from human manipulability to the robot one.  
With this transformation any desired set of manipulability ellipsoids can be adapted in real-time to the manipulability domain of a robotic system. For the human-to-robot transfer this can be achieved by recording human motion with a motion-capture system, expressing it in a musculoskeletal model of the human arm, e.g. the rHuMan framework \cite{figueredo2020human}, and then transferring the manipulability information herein to the robot domain.
%

The proposed domain adaptation method can be learned offline, prior to any training, and 
takes inspiration from the field of EEG signal processing 
and computer vision.
It perceives the domain adaptation as a point cloud registration problem on the SPD manifold. The method uses an analytic expression for the parallel transport between two domains as well as an iterative refinement with a manifold-aware ICP algorithm.
%
%
%
%
%
It is worth highlighting that the proposed solution can straightforwardly be applied to the 
%
transfer of additional features that are given in the form of SPD matrices, for instance stiffness and inertia.

\section{Related Work}\label{sec:related_work}

To realize such a framework that integrates and transfers task-oriented manipulation postures from humans to robots, this work builds on and contributes to the literature of (i)  learning by demonstration, particularly, to recent works concerning posture learning and (ii) Riemannian-aware domain adaptation -- as can be found in classification of EEG-based signals
by means of covariance matrices.



\textit{Robotics-oriented manipulability learning} 
investigates how to learn a time-varying manipulability profile -- embedding posture-dependent requirements such as preferred directions for motion/forces -- from a set of teacher demonstrations to be later deployed in the learner system. Early studies on the topic explored optimization methods, for instance to achieve the desired manipulability volume
\cite{lee2016humanoid}
or voxelized structures that discretize the workspace \cite{Vahrenkamp2012,figueredo2020human}. 

These approaches however neglect the topological structure of manipulability ellipsoids. 
Taking inspiration on the quaternion formulations for dynamic motion primitives (DMPs), 
recent works explored the Riemannian manifold of SPD matrices to ensure learning with respect to the proper group topology, e.g., \cite{rozo2017,geometryawaredmps,jaquier2018,jaquier}.   
%

In \cite{rozo2017}, Rozo et al. learn manipulability ellipsoids from multiple demonstrations using a Gaussian Mixture Model (GMM) on the SPD manifold, and deploy them at the learner via a redundancy-resolution controller scheme. 
Another LbD approach based on geometry-aware DMPs has been suggested by Fares et al. \cite{geometryawaredmps}. In this work, the manifold of SPD matrices is employed to embed both stiffness and manipulability profiles in different scenarios using a DMPs-based formulation. Furthermore, 
the proposed solution explores different teacher-learner systems -- yet without any domain adaptation to system capabilities, i.e., it is assumed that the learner can reproduce the learned profile efficiently.  
%
%
Jaquier et al. \cite{jaquier2018} propose a manipulability tracking framework on the SPD-manifold 
based on an extension of the classical inverse kinematics formulation. They exploit a tensor-based Jacobian to  map the changes in joint velocities to changes in manipulability.  

Overall, \cite{geometryawaredmps,jaquier2018,rozo2017} confirm that improved results are achieved when employing manifold-aware methods compared to methods that neglect the underlying geometry and are based on Euclidean space. 
Integrating previous findings, a complete geometry-aware manipulability learning, tracking, and transfer framework is proposed in \cite{jaquier}. All steps from learning to execution
are performed on the SPD manifold. 
This work is extended to address for the first time a transfer from human demonstrated manipulability to a robotic system in \cite{jaquieranalysis}. 
%
However, the results still consider direct learning and reproduction without concerns about the differences in topological structure of the systems -- and the domain of feasible manipulabilities within the manifold of SPD matrices. 
%

The idea of expressing desired manipulabilities as a function of the student and teacher capabilities for proper adaptation has been already proposed in \cite{jaquieranalysis,jaquier}.  
Yet, to the best of our knowledge, such an adaptation method has never been addressed and remains an open problem to this date. 

\textit{Riemannian-aware domain adaptation} 
has shown, on the other end, promising results in the fields of computer vision and medical data processing. Such methods particularly are employed when data is present in the form of covariance matrices and descriptors. Applications range from object detection \cite{manifoldobjectdetection} and image synthesis \cite{ganmanifold} to medical image segmentation \cite{manifoldimagesegmentation} and classification \cite{zaninitransfer}.   
%

In terms of domain adaptation on the SPD manifold, one of the first works is \cite{wang2008manifold} which focuses on transfer learning for language translation and is based on manifold alignment through Procrustes analysis. More 
recently, Cui et al. \cite{manifoldadapt} proposed a covariance shift for domain-adjusted image classification. Indeed, when it comes to manifold-aware domain adaptation classification has been 
one of the central topics addressed. Particularly, it has been well-studied in the context of brain-computer-interface (BCI) analysis where EEG information is embedded within covariance matrices. The relevance to the medical field relies on the fact that the acquired data consist of measurements from multiple subjects and several sessions, and thus do not live in the same region of the manifold -- in fact, transformations can happen even within data of the same subject as an effect of varying external conditions \cite{yairot}. 
This leads to the urgent necessity for domain adaptation on the manifold, which is similar to the problem addressed herein.  
%
%
%

Among the most successful methods, it is worth mentioning \cite{braineeg,yairot,zaninitransfer}.  
In \cite{zaninitransfer}, a probabilistic GMM on the manifold is proposed for classification where improvements are obtained by the introduction of an additional affine transformation. Parallel to this study, O'Yair et al. \cite{yairot} suggest a more general parallel transport (PT) based formulation for the domain adaptation and transfer learning problem which takes the results within \cite{zaninitransfer} as a corollary of the proposed method. Taking inspiration from computer vision, 
Rodrigues et al. \cite{braineeg} suggest point cloud registration on the SPD manifold to adapt between different subjects -- also in the context of transfer learning for BCI. 

Our work builds on and extends the works from \cite{braineeg} and \cite{yairot} to design an efficient method for human-to-robot manipulability domain adaptation in the context of learning postures in accordance with differences among kinematic structures. 
%





\section{Preliminaries}


The ability to compute a quality index for manipulation tasks
has been widely used in robotics.  
Different metrics can be explored according to specific task objectives, as seen in further detail in \cite{2015_Patel_Tarek_JIRS,2017_Ajoudani_Tsagarakis_Bicchi}.  
The most widely-known technique is the geometric representation of manipulability ellipsoids -- as introduced in  \cite{1985_Yoshikawa_ICRA,1985_Yoshikawa_IJRR}. Let the joint space be of dimension $N$ and the task-space of dimension $D$.
A manipulability ellipsoid $\mymatrix{M}$ is defined as the locus of velocities or forces corresponding to a deformation of the unit-norm of joint space variables, i.e., $ || \dot{\textbf{q}} || {=} 1 $ where $\dot{\textbf{q}} \in \mathbb{R}^N$ is the vector of joint velocities.\footnote{With the virtual work principle, it is easy to show the duality between velocity and wrench quantities which leads to velocity and force ellipsoids with similar directional axes but opposite magnitudes \cite{1985_Yoshikawa_IJRR,2015_Patel_Tarek_JIRS}.} 
In other words, given the kinematic relationship $ \dot{\textbf{x}} = \textbf{J}(\textbf{q}) \dot{\textbf{q}} $ between task-velocities $\dot{\textbf{x}}$ and joint velocities $\dot{\textbf{q}}$, where
$  \textbf{J}(\textbf{q}) \in \mathbb{R}^{D \times N} $ is the geometric Jacobian of the system, the mapping from the unit-norm in joint space to Cartesian velocity is given by  
\begin{equation}
     || \dot{\textbf{q}} ||^2 _2 = \dot{\textbf{x}}^T ( \textbf{M} ) ^{-1} \dot{\textbf{x}},  
     \label{eq:manipulability}
\end{equation}
where $ \textbf{M} = \textbf{J} \textbf{J}^T $.  
The directions of the principal axes are given by the set of orthonormal output vectors of the singular value decomposition (SVD) of \textbf{J}, whilst their magnitude is given by the corresponding singular value. They depict the directions of velocity capabilities, i.e., the major axis is aligned with the direction the system can achieve greater velocity with the least joint motion. On the other end, this is the direction along which the system is more sensitive to perturbations -- due to the duality between force and velocity manipulabilities \cite{1985_Yoshikawa_IJRR,2015_Patel_Tarek_JIRS,1988_Chiu_IJRR}.  
%
In singular cases -- when one or more axes collapse to zero -- motion becomes unfeasible in that direction. 
For instance, when the human arm is fully extended, it cannot move in the direction of extension by only moving one joint -- motions in that direction would indeed require an infinite joint motion. 

From the formulation in \eqref{eq:manipulability}, it is easy to see that -- when the Jacobian is well-posed -- manipulability ellipsoids are SPD matrices by construction, i.e., $\mymatrix{M}\in \spdgroup{D}$, where
\begin{equation}
     \spdgroup{D} {:=} \left\{ \mymatrix{M} {\in} \realset ^{D \times D} |  \mymatrix{M}{=}\mymatrix{M}^T, \myvec{x}^T \mymatrix{M} \myvec{x} {>} 0, \forall  \myvec x {\in} \realset^{D}     \right\}.
    \label{eq:group of SPDs}
\end{equation}

\subsubsection{Manifold of SPD Matrices}
The set definition \eqref{eq:group of SPDs} describes points in a convex half-cone in the vector space of the real $D{\times}D$ symmetric matrices $\symgroup{D}$. The resulting topology forms a differentiable Riemannian manifold. Each point $\mymatrix M$ on a Riemannian manifold $\spdgroup{D}$ can be bijectively mapped to a vector space, i.e. the tangent space $ \tangentspace _{\mymatrix M } \spdgroup{D} $ at $\mymatrix M$, and it is equipped with an inner product    
\begin{equation}
    \langle \mymatrix L _1, \mymatrix L_2 \rangle _ {\mymatrix M} 
            =  \langle \mymatrix M ^{-\frac{1}{2}}\mymatrix{L}_1 \mymatrix{M}^{-\frac{1}{2}},              
                        \mymatrix{M}^{-\frac{1}{2}}\mymatrix{L}_2 \mymatrix{M}^{-\frac{1}{2}} 
               \rangle_F , 
    \label{eq:inner product 1}
\end{equation}
which is equivalent to 
$
\langle \mymatrix{L}_1,\mymatrix{L}_2 \rangle_{\mymatrix{M} 
= 
\langle \mymatrix{M}^{-1}\mymatrix{L}_1 \mymatrix{M}^{-1} \mymatrix{L}_2\rangle}, 
$
where  $ \mymatrix{L}_1, \mymatrix{L}_2 \in \tangentspace{} _{\mymatrix{M} } \spdgroup{D} $.
A particular feature of the $\spdmanifold{}$ manifold is the existence of a unique geodesic between two points \cite{minh,yairot} -- which is defined based on the inner product. 
The arc length of the geodesic equals the Riemannian distance, which is also unique, but allows for affine invariance. 
That is, given the distance function $d$ between two points $\mymatrix{M}_1$ and $\mymatrix{M}_2$ on $ \spdgroup{D}$, then 
\begin{equation}
    d(\mymatrix{M}_1, \mymatrix{M}_2) = d(\mymatrix{M}_1, \mymatrix{A}\mymatrix{M}_2 \mymatrix{A}^T),
    \label{eq:distance affine}
\end{equation}
where $ \mymatrix{A} {\in} \realset^{D{\times}D} $ is an invertible matrix.
The affine invariant Riemannian distance between $\mymatrix{M}_1$ and $\mymatrix{M}_2$ on $ \spdgroup{D}$ is given by (see further details in \cite{yairot,braineeg})  
\begin{equation}
    d(\mymatrix{M}_1,\mymatrix{M}_2) = 
            ||\log(\mymatrix{M}_1^{-\frac{1}{2}}\mymatrix{M}_2\mymatrix{M}_1^{-\frac{1}{2}})||_F 
            = \sum_i \log^2(\lambda_i),  
    \label{eq:distance}
\end{equation}
where $\log$ is the matrix logarithm and $\lambda_i$ is the $i$-th eigenvalue of $\mymatrix{M}_1^{-\frac{1}{2}}\mymatrix{M}_2\mymatrix{M}_1^{-\frac{1}{2}}$.   
It equals the geodesic length between $\mymatrix{M}_1$ and $\mymatrix{M}_2$. 
Exploring the distance between points on the manifold, we can further define the geometric mean of multiple points. The geometric mean of a set of points/matrices $\{\mymatrix{M}_1, ..., \mymatrix{M}_N\} \in \spdgroup{D}$ is defined as
\begin{equation}
     \overline{\mymatrix M} = 
     \arg \min_{\mymatrix{X}\in \spdgroup{D}} 
     \sum_i
     d(\mymatrix{M}_i, \mymatrix{X}).
    \label{eq:geometric mean}
\end{equation} 
For $N=2$ the solution of the optimization problem \eqref{eq:geometric mean} exists in closed-form. In the case of $N>2$ typically an iterative algorithm is used to find $\overline{\mymatrix M} $ \cite{yairot}. 

Finally, for the remainder of this work, we will explore logarithmic and exponential maps which project a point $\mymatrix M {\in }\spdgroup{D}$ to the tangent space $ \tangentspace _{ {\mymatrix P} } \spdgroup{D}  $ at any $\mymatrix{P} {\in} \spdgroup{D}$, 
and from the point $ \mymatrix{L} {\in} \tangentspace _{ \mymatrix P } \spdgroup{D} $ back to the manifold, 
\begin{align}
    \mymatrix{L} &= \log _{ \mymatrix{P} }\left( \mymatrix M  \right)
                 = \mymatrix{P}^{\tfrac{1}{2}}     
                        \log \left( 
                                \mymatrix{P}^{{-}\tfrac{1}{2}}  
                                        \mymatrix{M}   
                                \mymatrix{P}^{{-}\tfrac{1}{2}}               
                             \right)
                    \mymatrix{P}^{\tfrac{1}{2}},
\\     
    \mymatrix{M} &= \exp _{ \mymatrix{P} }\left( \mymatrix L  \right)
                 = \mymatrix{P}^{\tfrac{1}{2}}     
                        \exp \left( 
                                \mymatrix{P}^{{-}\tfrac{1}{2}}  
                                        \mymatrix{L}   
                                \mymatrix{P}^{{-}\tfrac{1}{2}}               
                             \right)
                    \mymatrix{P}^{\tfrac{1}{2}}.  
\end{align}
Elements on the tangent space $\tangentspace _{ \mymatrix P } \spdgroup{D} $ are in $\symgroup{D}$.

\section{Manipulability Domain Transfer}

\begin{algorithm}[t]
		\caption{Manifold-Aware ICP with PT Initialization}
        \label{alg:icp}
		
        \begin{algorithmic}
		\Procedure{}{}
            \State  $\overline{\mymatrix{S}},\overline{\mymatrix{T}} \gets$ initialize geometric means 
            \State  $s \gets$ initialize dispersion factor 
            \State$ { \mymatrix{E}} \gets (\overline{\mymatrix{S}}(\overline{\mymatrix{T}})^{-1})^\frac{1}{2}$ \Comment{PT}
           
            \For{all $\mymatrix{T}_i\in \mathcal{T}$}
                \State $\mymatrix{T}_i \gets \mymatrix{E}\mymatrix{T}_i\mymatrix{E}^T$ \Comment{PT}
            \EndFor
             \For{all $\mymatrix{T}_i\in\mathcal{T}, \mymatrix{S}_i\in\mathcal{S}$} 
                \State $\mymatrix{T}_i \gets$ $\overline{\mymatrix{T}}^{-\frac{1}{2}}\mymatrix{T}_i\overline{\mymatrix{T}}^{-\frac{1}{2}}$; \Comment{ICP Step 1}
                \State $\mymatrix{S}_i \gets$
                $\overline{\mymatrix{S}}^{-\frac{1}{2}}\mymatrix{S}_i\overline{\mymatrix{S}}^{-\frac{1}{2}}$
                \State $\mymatrix{T}_i \gets$ $\mymatrix{T}_i^s$; \Comment{ICP Step 2}
            \EndFor 
            \For{iter} \Comment{ICP Step 3}
                \State  $w,\text{point pairs} \gets$ correspondence matching 
                \State  $\mymatrix{R} \gets$ rotation optimization 
                \State  $\mymatrix{T}_i \gets$ $\mymatrix{R}\mymatrix{T}_i\mymatrix{R}^T$
            \EndFor
            \For{$\mymatrix{T}_i\in\mathcal{T}$} \Comment{ICP Step 4}
                \State $\mymatrix{T}_i \gets$ $\overline{\mymatrix{S}}^{\frac{1}{2}}\mymatrix{T}_i\overline{\mymatrix{S}}^{\frac{1}{2}}$
            \EndFor
\EndProcedure
	\end{algorithmic}
	\end{algorithm}

The adaptation of a task-dependent manipulability profile to the capabilities of a kinematically different system gives rise to a domain adaptation problem.
Every kinematic model can generate a characteristic set of manipulability matrices -- which we refer to as the \textit{manipulability domain} of that system. 
For the transfer between two domains, e.g. when adjusting a human-generated manipulability profile to the robot's capabilities, a transfer function $ f : \mathcal{T} \rightarrow \mathcal{S}$ 
from the teacher domain $\mathcal{T}$ to the student domain $\mathcal{S}$, is sought. All $\mymatrix{T}_i \in \mathcal{T}$ and $\mymatrix{S}_i \in \mathcal{S}$ are elements of $ \spdmanifold{}$ 
which on the manifold are represented by points. Therefore each manipulability domain is represented as a set of points on the $\spdmanifold{}$ manifold.

This perspective offers the possibility to view $\mathcal{S}$ and $\mathcal{T}$ as two point clouds on the $\spdmanifold{}$ manifold which inherently represent the characteristics of the two kinematic systems. The problem of finding a transformation between two or more point clouds is frequently found in computer vision and medical imaging and referred to as point cloud registration. 
It aims at aligning the \textit{target} $\mathcal{T}$ with the \textit{source} $\mathcal{S}$ s.t. $\mymatrix{S}_i = f(\mymatrix{T}_i) \quad \forall i \in {1,...,N}$, where $N$ is the number of points. In our setting the target $\mathcal{T}$ is the human domain which we aim to align with the robot source domain $\mathcal{S}$.

Undoubtedly one of the most popular point cloud registration algorithms is the Iterative Closest Points algorithm (ICP). It is based on the assumption that $\mathcal{T}$ can be aligned with $\mathcal{S}$ in Euclidean space by a rigid transformation, i.e. a translation $\mymatrix{B}$, a rotation $\mymatrix{R}$, and a scaling $s$, of the target point cloud. It furthermore assumes that for each $\mymatrix{T}_i$ there is a corresponding $\mymatrix{S}_i$. Based on these point correspondences the ICP iteratively finds the optimal transformation parameters of the following objective:
\begin{equation}
    J(\mymatrix{R},\mymatrix{B},s) = \underset{\mymatrix{R},\mymatrix{B},s}{\text{arg\,min}}\quad {\sum}_{i=1}^{N} || d(\mymatrix{S}_i, s\mymatrix{R}\mymatrix{T}_i+\mymatrix{B})||\label{eq:icpobjective}
\end{equation} 


For the adaptation of manipulability domains on the $\spdmanifold{}$ manifold, we modify the standard ICP formulation such that it identifies the optimal rotation, scaling and translation of a domain on the $\spdmanifold{}$ manifold. 

\subsection{Riemannian-ICP Algorithm}

In the following, we give a detailed description of the novel ICP algorithm for manipulability domain adaptation on the $\spdmanifold{}$ manifold. Our work builds on the EEG classification method by Rodrigues et al. \cite{braineeg} and the Parallel Transport method by O'Yair et al. \cite{yairot}.
 We assume $|\mathcal{S}| = |\mathcal{T}| = N$.

\noindent \textit{Step 1: Recentering to Identity:} First, the translation of the source $\mathcal{S}$ and the target $\mathcal{T}$  is removed by recentering both at the Identity $\mymatrix{I} \in \spdmanifold{}$ via
\begin{equation}
    \sourcepointtilde = \sourcemean^{-\frac{1}{2}}\sourcepoint\sourcemean^{-\frac{1}{2}} \quad \forall i \in 1...N \label{eq:recenters}
\end{equation} 
\begin{equation}
    \targetpointtilde = \targetmean^{-\frac{1}{2}}\targetpoint\targetmean^{-\frac{1}{2}} \quad \forall i \in 1...N \label{eq:recentert}
\end{equation}
where $\sourcemean$ and $\targetmean$ are the geometric means \eqref{eq:geometric mean} of $\mathcal{S}$ and $\mathcal{T}$, respectively. 
It is important to highlight that the recentering happens within the manifold and its specific domain. The two recentered point clouds are referred to as $\mathcal{S}^{rct}$ and $\mathcal{T}^{rct}$ in the following.

\noindent \textit{Step 2: Dispersion Adjustment:} In the second step the dispersion around the mean of $\mathcal{T}^{rct}$ is adjusted such that it matches the dispersion of $\mathcal{S}^{rct}$. This step is equivalent to the scaling step in the Euclidean-based ICP. To this end a scaling of $\mathcal{T}^{rct}$ is performed with the matrix power:
\begin{equation}
\targetpoint^{scl} = (\targetpointtilde)^{\negthickspace\frac{c_S}{c_T} } \quad \forall i \in 1...N
\label{eq:dispersionadjust}
\end{equation} 
where 
$c_S$ and $c_T$ are the dispersions of the respective point clouds, 
\begin{align}
    c_S &= \frac{1}{N}{\sum}_i^N d(\sourcepointtilde - \mymatrix{I}) ,\label{eq:disps}
\\
    c_T &= \frac{1}{N}{\sum}_i^N d(\targetpointtilde - \mymatrix{I}) . \label{eq:dispt}
\end{align}
This operation moves the target points $\targetpointtilde$ along the geodesics connecting them with the identity $\mymatrix{I}$ such that the area covered by the point cloud $\mathcal{T}^{rct}$ on the manifold matches the area of $\mathcal{S}^{rct}$.

\noindent \textit{Step 3: Rotation Optimization:} In the third step the optimal rotation matrix $\mymatrix{R}$ is found, where $\mymatrix{R}^T\mymatrix{R}=\mymatrix{I}$ and $\mymatrix{R} \in {\sogroup}$. This is achieved in an iterative fashion until a stopping criterion is reached: In every iteration, first, the correspondences are matched and weights $w_i$ are determined based on the current estimate. Then $\mymatrix{R}$ is estimated by optimizing 
\begin{equation}
\underset{\mymatrix{R}^T\mymatrix{R}=\mymatrix{I}}{\text{min}}\quad \sum_{i=1}^{N} w_i d^2(\sourcepoint^{rct}, \mymatrix{R}\targetpoint^{scl}\mymatrix{R}^T). 
\label{eq:rotopt}    
\end{equation}

Prior to the next iteration the rotation $\mymatrix{R}$ is applied to the scaled target point cloud with
\begin{equation}
    \mymatrix{T}_i^{rt} = \mymatrix{R}\mymatrix{T}_i^{scl}\mymatrix{R}^T \quad \forall i=1,...,N.
    \label{eq:rotapply} 
\end{equation}
Our implementation is realized in Python with the Pymanopt framework \cite{pymanopt} which provides optimization functionalities on the $\spdmanifold{}$ and $\sogroup$ manifolds.

\noindent \textit{Step 4: Translation Adjustment:} In the last step the rotated target point cloud $\mathcal{T}^{rt}$ is translated such that their means coincide:
\begin{equation} 
\targetpoint^{tl} = 
    \sourcemean^{\frac{1}{2}}
            \targetpoint^{rt}
    \sourcemean^{\frac{1}{2}} \quad \forall i=1,...,N.
\label{eq:translationadjust}
\end{equation}
Now, $\mathcal{T}^{tl}$ is optimally aligned with $\mathcal{S}$ and the transformation parameters to transfer the teacher manipulability domain $\mathcal{T}$ to the student domain $\mathcal{S}$ have been found. The algorithm is summarized in Algorithm \ref{alg:icp}. A newly observed set of manipulabilities $\mathcal{T}_{new}$ generated by the teacher can thus be efficiently transferred to the student domain by applying the transformation $\mathcal{S}_{new} = f(\mathcal{T}_{new})$ with \eqref{eq:recentert}, \eqref{eq:dispersionadjust}, \eqref{eq:rotapply} and \eqref{eq:translationadjust}. It is worth highlighting that the estimation of the parameters only needs to be done once and can be achieved in an offline manner, while the adaptation of newly observed manipulabilities is then efficiently achieved online.

\subsection{Correspondence Matching}

The ICP algorithm is based on the assumption that for each $\targetpoint$ there exists a corresponding $\sourcepoint$. When prior knowledge of the point correspondences is not given a matching algorithm has to be employed before conducting the ICP. Since the introduction of the original ICP algorithm by Besl and McKay \cite{besl1992icp} and Chen and Medioni \cite{chen1992icp} various techniques have been proposed for this problem. 
Different matching heuristics have been introduced such as the point-to-point distance \cite{besl1992icp}, color \cite{weik1997color} and image feature descriptor compatibility \cite{kamencay2019featuredescriptor} or point-to-plane distance \cite{chen1992icp}.

For the matching of manipulability ellipsoids -- represented as points on $ \spdmanifold{} $ -- and determination of the weights $w_i$ in \eqref{eq:rotopt} we propose a heuristic based on intrinsic elliptic geometric features.
Two ellipsoids match each other when (i) their minor  and (ii) principal axes are aligned and when (iii) their singularity indices and (iv) volumes are identical. Therefore we match each $\targetpoint$ with the $\sourcepoint$ that maximizes the weighting function 
\begin{align}
w_i
    & = {{|\myvec{v}_1^{\targetpoint}\cdot \myvec{v}_1^{\sourcepoint}|} +                            {|\myvec{v}_n^{\targetpoint}\cdot \myvec{v}_n^{\sourcepoint}|}}\label{eq:weights}\\
    & + { e^{-|p^{\targetpoint}-p^{\sourcepoint}|} +             {e^{-|vol^{\targetpoint}-vol^{\sourcepoint}|}}}\nonumber 
\end{align} 
where $\myvec{v}$ are the eigenvectors corresponding to the eigenvalues $\lambda_1 \leq ... \leq \lambda_n$ of the respective element. The inverse condition number $p = \lambda_n/\lambda_1$ indicates the closeness to singularity \cite{togai1986application}, i.e. proportion between the major and minor axes, while $vol$ refers to the ellipsoid volume.
We perform the matching process in every rotation optimization iteration so that correspondences can be established based on the current alignment estimate. 

In the following, we will validate the proposed manipulability matching heuristic on a toy dataset. For the source domain $\mathcal{S}$ we randomly sample $100$ manipulabilities of the 7-DoF Panda model. The target domain $\mathcal{T}$ is generated by a rigid transform of $\mathcal{S}$ on $\spdmanifold{}$ with randomly generated rotation, scaling, and translation. We then match the points with (\ref{eq:weights}) and recover the transformation parameters with the Riemannian-ICP. 
In a first variant all 100, while in a second variant only the 12 most singular samples  -- according to their singularity indices -- of both datasets, are considered for the matching and alignment process.

\begin{figure}[t]
    \centering
       \includegraphics[width=1\columnwidth]{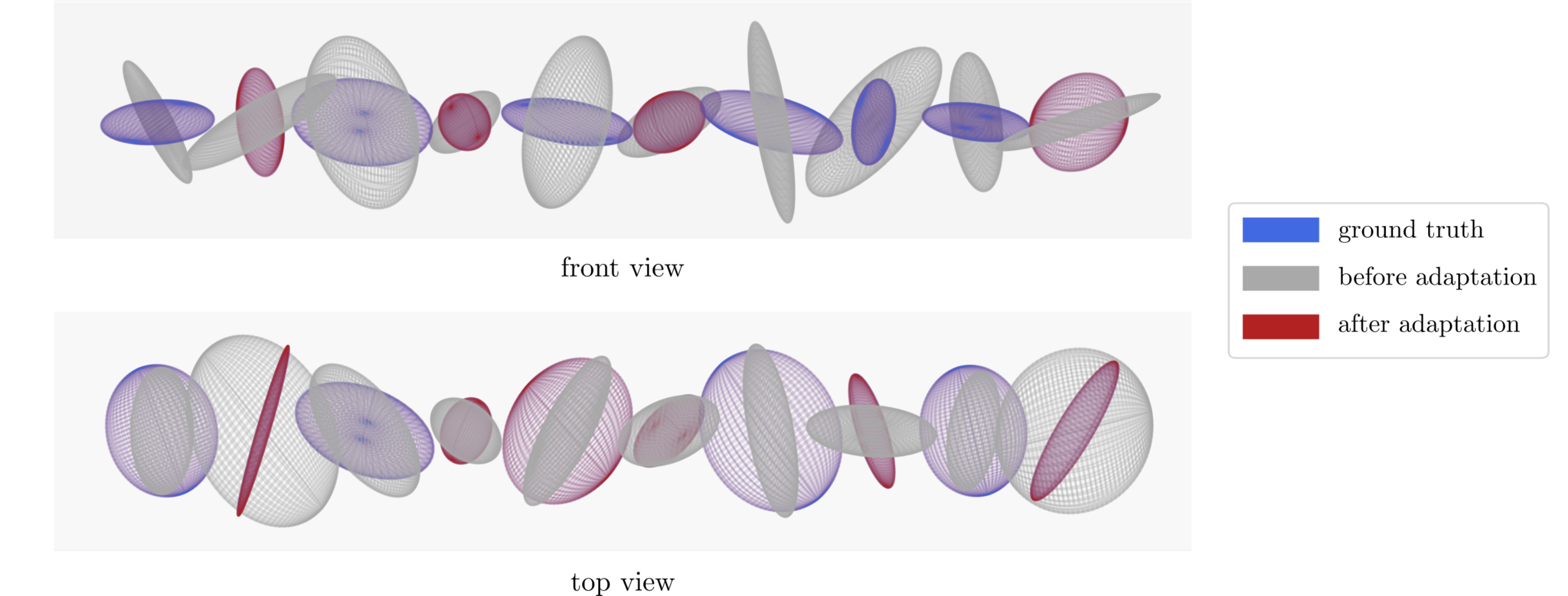}
    \caption{Toy data validation set (7-DoF) before and after domain adaptation. Ellipsoids are shown from top and front view. After adaptation the major and minor axes align well with the ground truth and the ellipsoids coincide.}
    \label{fig:toydata}
\end{figure}

The results for the transfer of $10$ new randomly sampled points to the source domain are presented in Table \ref{tab:toydata} and Figure \ref{fig:toydata}. They show that in both cases the rigid transformation has been successfully recovered. Table \ref{tab:toydata} shows that When considering $100$ points the convergence is faster and the resulting manipulabilities are closer to the ground truth. The source and target ellipsoids in Figure \ref{fig:toydata} confirm the successful alignment.


Overall it was shown that the  proposed heuristic is capable of matching corresponding manipulability ellipsoids correctly and that the Riemannian-ICP can successfully perform manipulability domain adaptation in the presence of a perfect rigid transformation between $\mathcal{S}$ and $\mathcal{T}$ on the $\spdmanifold{}$ manifold.

    

\begin{table}[t]
	\caption{Toy dataset (7-DoF): RMSE and iterations until convergence for domain adaptation of 10 newly sampled points of the target domain.}
	\label{tab:toydata}
	\centering
	\begin{tabular}{cc|c}
		\hline
		\hline
		Training Samples 
		& iterations& RMSE\\
		\hline
        12 most singular of 100 & 28& 0.095\\
        100 & 22&0.042\\
		\hline
		\hline
	\end{tabular}
\end{table}


\subsection{Initialization with Parallel Transport}

It is well-known that the ICP often converges to a local minimum which leads to bad performance. Providing a good initial alignment estimate for the ICP can draw the optimization process towards the global optimum \cite{hugli1997initial, dorai1997initial, stein1992initial}. 
To find such an initial estimate for the Riemannian-ICP we employ the parallel transport (PT) based alignment method of O'Yair \cite{yairot} as an additional step prior to applying the ICP algorithm. 
We apply the map $\Psi(\targetpoint) \widehat{=} \mymatrix{E}\targetpoint\mymatrix{E}^T$ individually to every target point with $ \mymatrix{E} = (\overline{\mymatrix{S}}(\overline{\mymatrix{T}})^{-1})^\frac{1}{2}$. It is a closed form solution for the parallel transport operation from $\mathcal{T}_{\targetmean{}}\spdmanifold{}$ to $\mathcal{T}_{\sourcemean{}}\spdmanifold{}$.
With $\Psi$, the target points are adapted to the source
domain while keeping their relative position to the mean, i.e. while preserving their
geometric relationship in a manifold-aware manner.

\section{Experimental Evaluation}
This section presents a variety of simulated experiments deploying the proposed manipulability domain adaptation between different systems. With the experiments we aim to verify the proposed domain adaptation scheme and to explore how the size and type of training data, weights and the PT influence the results.

All scenarios are implemented using Python 3.6.9 and the 
DQ-Robotics library \cite{Adorno2019_dqrobotics} on a standard workstation\footnote{ Ubuntu LTS with a Intel® Core™ i5-7200U CPU @ 2.50GHz × 4}.
%
For simplicity throughout all experiments, we only consider the translational part of the robot Jacobian, resulting in manipulabilities of size $3\times 3$.
For the performance evaluation, a dispersion normalized version of the Riemannian RMSE is used:
\begin{equation}
    RMSE = \frac{1}{c}\sqrt{\tfrac{1}{N}{\sum}_{i=1}^{N}d(\mymatrix{S}_i,\mymatrix{T}_i)^2} \label{eq:rmse}
\end{equation}
where $c$ is the dispersion of the dataset after the mapping. 

The proposed algorithm is highly sensitive to the initialization value of the rotation optimization routine in \eqref{eq:rotopt}. Therefore a multi-point random initialization is used throughout the experiments. 

Finally, for all models deployed in this Section, we create two training datasets for the optimization with the manifold-aware ICP: A random dataset by extracting manipulabilities from uniform joint angle samples satisfying joint constraints and a trajectory-based dataset with several manipulability points following predefined geodesics.

\subsection{Manipulability Domain Adaptation with Known Heuristics}
\begin{figure}
    \centering
    \includegraphics[width=\columnwidth]{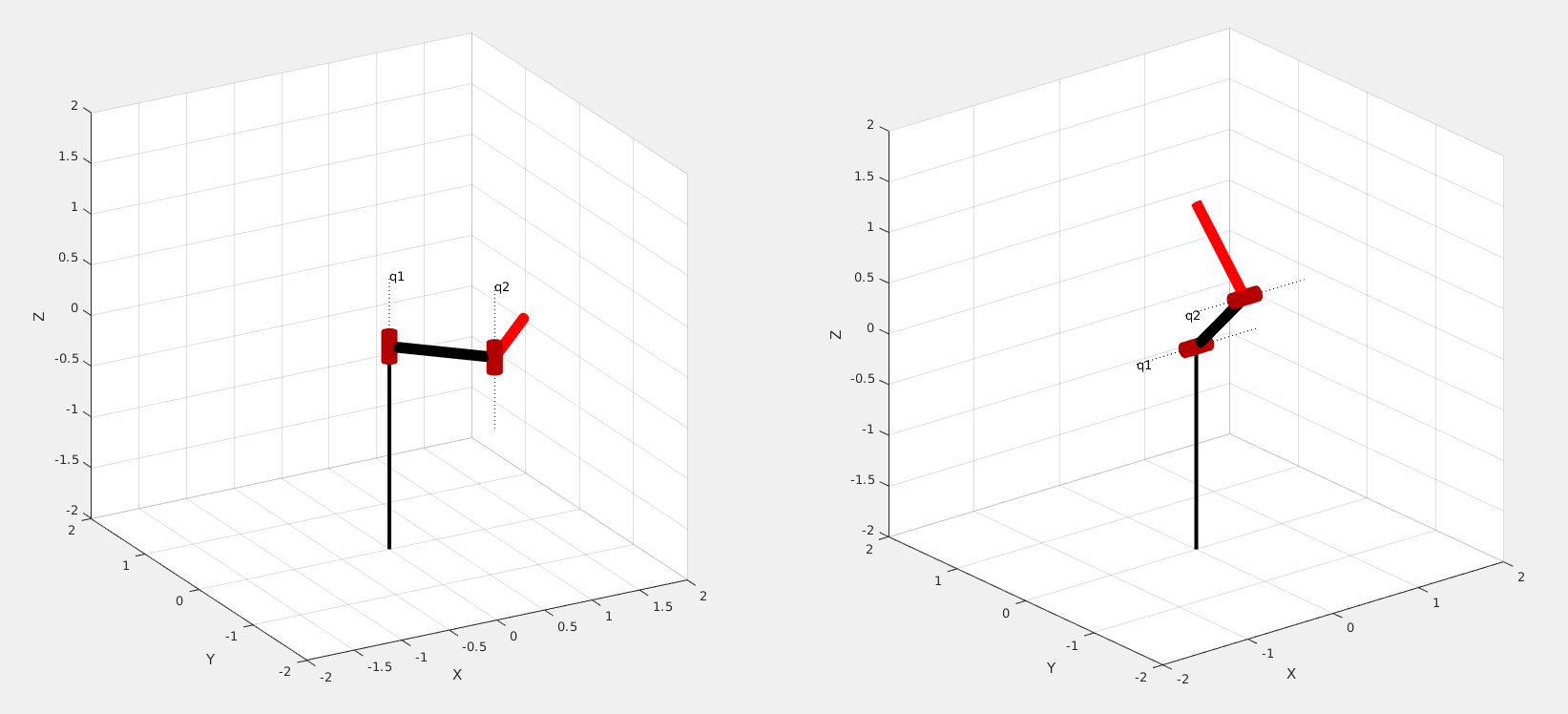}
    \caption{The 2-DoF serial manipulators, left: horizontal (teacher), right: vertical (student)}
    \label{fig:figure_2dofs}
\end{figure}

In the first experiment, domain adaptation between two 2-DoF serial manipulators (see Fig. \ref{fig:figure_2dofs}) is performed.
In this scenario, the trajectory-based dataset consists of $20$ trajectories during which one joint is fixed at varying angles while the other performs a $180 ^{\circ}$ rotation\footnote{Any zero eigenvalue encountered during data generation is manually replaced by $1e-4$ to project it onto $\spdmanifold{}$. This holds for all simulated data in our work.}. 

As the serial manipulators only differ in their orientation in the world space, i.e. a rotation of $90^{\circ}$, the respective manipulabilities do too -- according to our heuristic. The evaluation of the proposed Riemannian manifold-aware ICP is based on $3$ further trajectories that are generated for each domain, namely \textit{M-Eval-1}, \textit{M-Eval-2}, and \textit{M-Eval-3}. Table~\ref{tab:table2dofs} shows the RMSE of the 3 trajectories after $100$ optimization iterations for different configurations such as the size and type of the training dataset and weighting settings. For the smaller datasets, 
we subsample the main dataset by taking every $n$-th point. 
Furthermore, we evaluate how the parallel transport (PT) initialization affects the results.

\begin{figure}[t]
    \centering
       \includegraphics[width=\columnwidth]{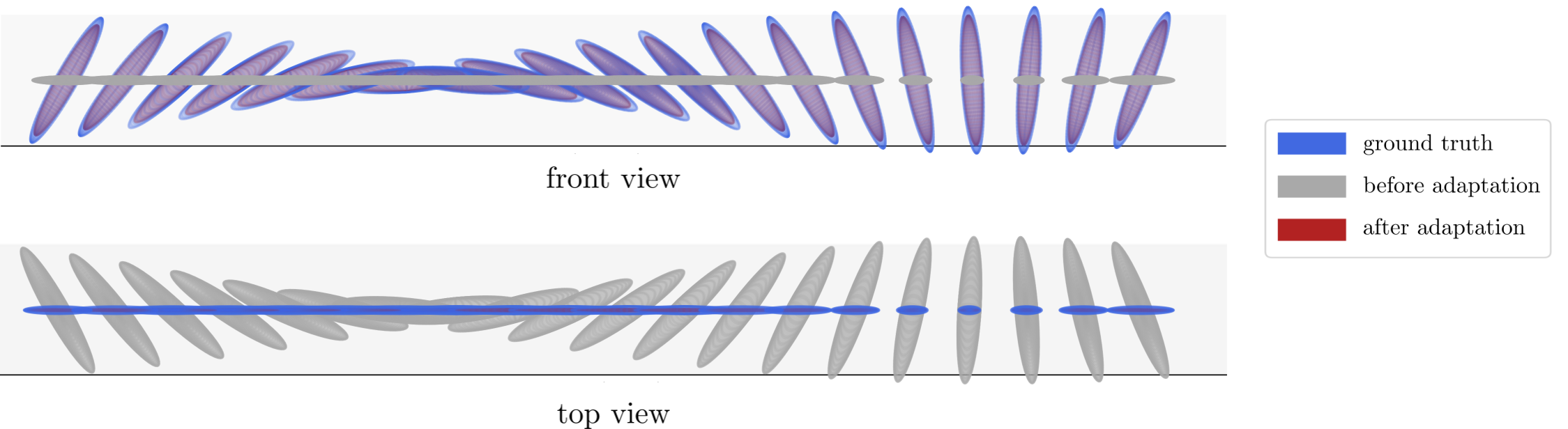}
       \label{fig:table2_mapped_manip18_front}
    
    \caption{The \textit{M-Eval-1} validation set before and after domain adaptation, viewed from top and front. After adaptation the major and minor axes are well aligned with the ground truth, only the volume is slightly smaller.
    }
    \label{fig:manips2dof}
\end{figure}

The results reveal that the PT initialization increases performance significantly and that a weighting of $w^3$ -- where $w$ is as presented in (\ref{eq:weights}) -- is beneficial.
The results furthermore show that more training points do not necessarily lead to better results. Figure~\ref{fig:manips2dof} visualizes the resulting ellipsoids for the \textit{M-Eval-1} trajectory together with the \textit{PT + ICP + $w^3$} variant and 100 training samples and confirms the successful domain adaptation between the 2-DoF robots. Still, the transformed manipulabilities are slightly smaller than the ground truth as expected from Table \ref{tab:table2dofs}.

\begin{table}[h!]
	\caption{Resulting RMSE for 2-DoF generated trajectories (upper half) and random data (lower half)}
	\label{tab:table2dofs}
	\centering
	\begin{tabular}{cc|ccc}
		\hline
		\hline
		No. samples & method &M-Eval-1&M-Eval-2&M-Eval-3\\
		\hline
		400 & PT only&0.874&0.783&0.782\\ 
		200 & ICP & 1.058&0.827&0.829\\ 
		100 & ICP + $w$&0.913&0.725&0.726\\ 
		100 & ICP + $w^2$ &0.933&0.733&0.733\\ 
		100 & ICP + $w^3$ &0.710&0.606&0.605\\ 
		100 & PT + ICP + $w^3$ &\textbf{0.362}&\textbf{0.410}&0.419\\ 
		50 & ICP only + $w^3$&0.714&0.618&0.606\\ 
		50 & PT + ICP + $w^3$&0.703&0.594&0.912\\ 
		25 & ICP only + $w^3$&0.548&0.547&0.453\\ 
		25 & PT + ICP + $w^3$&0.440&0.497&\textbf{0.408}\\ 
		\hline
		500 & PT only&0.875&0.774&0.790\\ 
		100 & ICP only + $w^3$&0.864&0.715&0.708\\ 
		100 & PT + ICP + $w^3$&0.864&0.736&0.726\\
		50 & ICP only + $w^3$ &0.648&0.550&0.547\\ 
		50 & PT + ICP + $w^3$&0.649&0.540&0.538\\
		25 & ICP only + $w^3$&0.677&0.545&0.554\\ 
		25 & PT + ICP + $w^3$&\textbf{0.586}&\textbf{0.492}&\textbf{0.496}\\
		\hline\hline
	\end{tabular}
\end{table}

\subsection{Human-to-Robot Manipulability Domain Transfer}

In the second experiment we aim to evaluate the proposed method for the domain adaptation between a 7-DoF human arm model and the 7-DoF Panda robot.
For the human arm model, we use the rHuMan toolbox from \cite{figueredo2020human} which is based on the biomechanics-aware model of \cite{Saul2015}, whilst 
the robot model relies on the standard kinematics of the Panda-arm. 

For both models, we simulate trajectories of moving the extended arm either upwards or sideways at different shoulder elevation/rotation angles. The trajectories are chosen such that they resemble frequent arm movements -- like bending the elbow or rotating the shoulder -- and do not violate any joint constraints. Here, the trajectory-based dataset consists of $8$ trajectories and $4$ trajectories (\textit{H-Eval-1}, \textit{H-Eval-2}, \textit{H-Eval-3}, and \textit{H-Eval-4}) are used for the evaluation.

To the best of the author's knowledge, there exists no other solution for the domain adaptation problem in the context of manipulability learning. 
Existing strategies, e.g. \cite{jaquier,geometryawaredmps}, would most often simply copy one manipulability to the other without the proper understanding of their kinematic capabilities. 
Therefore, in this work we also devise a baseline approach based on a naive voxelized nearest neighbor search. 

The naive baseline is based on building voxelized structures of $5000$ manipulabilities within each domain (i.e., the human model and Panda-arm manipulability domains). 
A nearest neighbor search between both domains is performed for each data point. This yields for each data point an individual mapping from the human domain to the robot one. 
As soon as we acquire a new point in the human domain, we perform once again the nearest neighbor search -- yet within the same domain -- and explore the precomputed mapping to the robot domain. This is an elaborate procedure that takes several hours, and cannot be deployed for practical application. Nonetheless, it provides us with a rough baseline approach to evaluate the ICP results in terms of the RMSE. 

\begin{figure}[t]
    \centering
     
    
           \includegraphics[width=\columnwidth]{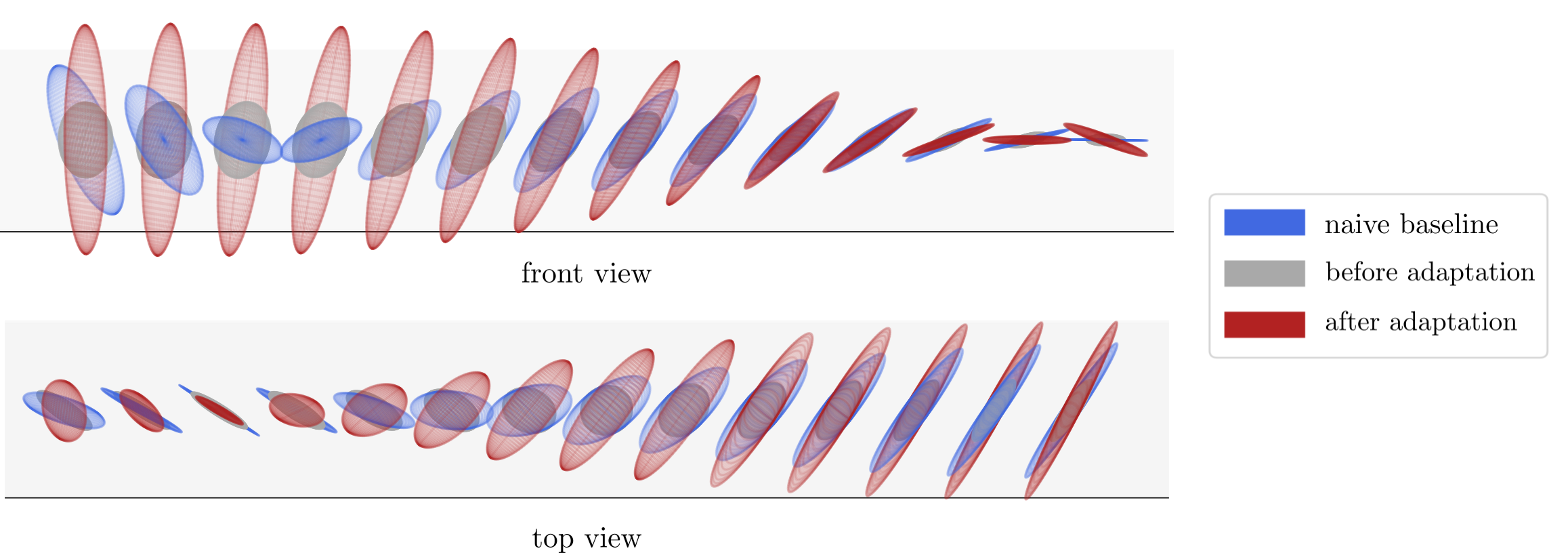}
    \caption{Manipulabilities of the human arm model before and after domain adaptation, viewed from top and front. Similarly to the human manipulabilities the adapted ones are transformed along a geodesic towards convergence.
    }
    \label{fig:humanmanips}
\end{figure}

The results of the human-to-robot domain transfer with the proposed manifold-aware ICP are shown in Table~\ref{tab:tablehr}. Once again it can be seen that the proposed PT initialization increases the performance. The resulting ellipsoids for the \textit{H-Eval-4} trajectory -- representing a vertical upward movement of the fully extended arm --, together with the \textit{PT + ICP} variant and the trajectory-based dataset with 25 samples, can be seen in Fig.~\ref{fig:humanmanips}. 
The manipulabilities are roughly transformed along a geodesic towards convergence -- similarly to the input data from the human model. Furthermore, it is worth noticing that the shape of the Panda manipulability domain converges for both the naive baseline approach and the proposed solution. 
For isotropic samples, Fig.~\ref{fig:humanmanips} shows a clear difference between the domain adapted manipulabilities from the naive baseline approach and the ICP method  -- likely because the matching heuristic captures the geometric features, particularly the orientation, for more singular samples better than for more isotropic samples. Notwithstanding, it is easy to observe that the transformation is less abrupt within our approach than with the naive baseline. 
%

\begin{figure}[t]
    \centering
    \includegraphics[width=1\columnwidth]{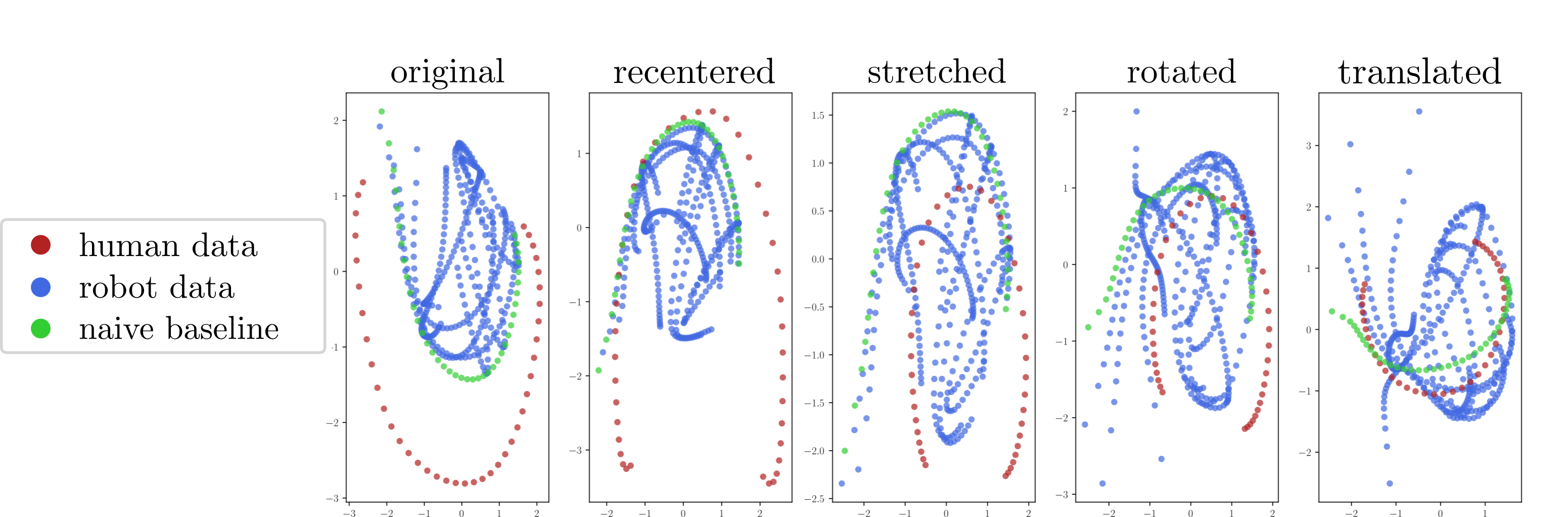}
    \caption{Domain adaption process of \textit{H-Eval-4}, visualized with diffusion maps \cite{diffusionmaps}. The data is transferred step-wisely by the parameters found with the manifold-ICP. After domain adaptation the human data aligns well with the robot training data.}
    \label{fig:mapping_up_60}
\end{figure}

Furthermore, the domain adaptation process of the \textit{H-Eval-4} trajectory with the transformation parameters identified by the proposed ICP is illustrated in Fig.~\ref{fig:mapping_up_60}.
It can be seen that after adaptation the trajectory is well aligned in between the other trajectories of the robot training data.

\begin{table}[b]
	\caption{Resulting RMSE for 7-DoF generated human and robot trajectories (upper half) and random data (lower half). RMSE is based on the naive baseline. $w^3$ is used and PT is applied in all cases.}
	\label{tab:tablehr}
	\centering
	\begin{tabular}{cc|cccc}
		\hline
		\hline
		\hspace{-6pt}Samples \hspace{-6pt}& Method & H-Eval-1 \hspace{-6pt} &H-Eval-2\hspace{-6pt}&H-Eval-3\hspace{-6pt}&H-Eval-4\hspace{-6pt}\\
		\hline
		400  & PT + ICP &\textbf{0.539}&\textbf{0.834}&\textbf{0.340}&0.297\\ 
		100 & ICP  &0.548&0.854&0.346&0.300\\ 
		100& PT + ICP &0.549&0.848&0.348&0.300\\ 
		25 & ICP  &0.578&0.884&0.376&0.297\\ 
		25& PT + ICP &0.576&0.888&0.375&\textbf{0.290}\\ 
		\hline
		100 & PT+ICP&0.602&0.827&0.340&0.377\\ 
		100 & ICP&\textbf{0.437}&0.699&\textbf{0.317}&0.363\\ 
		25 & PT+ICP&0.501&\textbf{0.698}&0.344&\textbf{0.370}\\ 
		25 & ICP&0.619&0.818&0.339&0.382\\ 
		\hline
		\hline
	\end{tabular}
\end{table}

\section{Conclusion and Future Work}\label{sec:conclusion}

In this work, a novel method for manipulability domain adaptation between two robots of different embodiments was proposed -- for the first time from the perspective of point cloud registration on the Riemannian manifold. It is based on a Riemannian manifold-aware ICP and parallel transport and finds a rigid transformation between both domains with respect to the underlying SPD topology of the manipulability matrices. In simulation experiments with 2-DoF and 7-DoF models we validated the approach.

In the future, we plan to conduct extensive experiments on human-generated motion data to open the door for a human-integrated PbD-based approach and furthermore extend the method to bimanual systems. Finally, a relaxation of the rigidity constraint is worth exploring and could possibly be realized with a manifold-aware coherent point drift algorithm.







%

\bibliography{bib.bib, Bibliography.bib, bibtex_THRI.bib, bibtexCoopAndOthers.bib, newreferences.bib}
\bibliographystyle{IEEEtran}



\end{document}